\documentclass[a4paper,english]{llncs}
\pdfoutput=1 
\usepackage{microtype}

\usepackage{amsmath}

\usepackage{icomma} 
\usepackage[final]{graphicx}
\usepackage{wrapfig} 
\usepackage{cmap}					
\usepackage[T2A]{fontenc}			
\usepackage[utf8]{inputenc}			
\usepackage[english]{babel}	

\usepackage{hyperref}
\usepackage[usenames,dvipsnames,svgnames,table,rgb]{xcolor}
\hypersetup{				
  unicode=true, 
  pdftitle={Semantic clustering of Russian web search results: possibilities and problems}, 
 pdfauthor={Andrey Kutuzov}, 
 pdfkeywords={computational semantics} {search engine results clustering} {graph theory}, 
  colorlinks=false, 	
  linkcolor=red, 
  citecolor=black, 
 filecolor=magenta, 
}
\usepackage{setspace} 
\usepackage{lastpage} 
\usepackage{soulutf8} 
\usepackage{array,tabularx,tabulary,booktabs} 
\usepackage{longtable} 
\usepackage{multirow} 

\begin{document}
\author{Andrey Kutuzov}
\institute{Mail.ru Group, National Research University Higher School of Economics}
\title{Semantic clustering of Russian web search results: possibilities and problems}

\maketitle

\begin{abstract}
The present paper deals with word sense induction from lexical co-occurrence graphs. We construct such graphs on large Russian corpora and then apply the data to cluster the results of Mail.ru search according to meanings in the query. We compare different methods of performing such clustering and different source corpora. Models of applying distributional semantics to big linguistic data are described.
\end{abstract}

\section{Introduction}\label{section:intro}

The presented paper deals with the problem of semantic clustering of search engine results page (SERP). The problem arises from the obvious fact that many user queries are ambiguous in some way. Thus, search engines strive to diversify their results and to present such results that are related to as many query interpretations as possible. For example, Google search for the Russian word \foreignlanguage{russian}{\textit{`максим'}} returns:
\begin{enumerate}
\item five results related to a popular singer,
\item two results for a magazine,
\item one result for \url{http://lib.ru}, Maxim Moshkow's electronic library,
\item one result for a proper name.
\end{enumerate}

However these results are not sorted by their meaning and are returned simply according to their relevance ranking, which for many of them seems to be almost equal. The obvious way to cluster the results is by the words their snippets share. Unfortunately, often snippets for results belonging to one query sense do not have a single content word in common (except for the query itself, which is useless). Cf. two snippets for the first query meaning from the example above:
\begin{enumerate}
\foreignlanguage{russian}{
\item \textit{`МакSим начинает самостоятельно заниматься своей карьерой, пишет новые песни. В этот период певица выступает как малобюджетный проект, ...'}
\item \textit{`МакSим презентовала видеоклип «Я буду жить», получивший широкую огласку еще до момента появления видео в сети.'}}
\end{enumerate}
They do not have a single common word, but still belong to one meaning (popular singer).

Moreover, snippets for different query senses can share some words. Cf. two snippets from the same search engine results page. They share the word \foreignlanguage{russian}{\textit{`автор'}} (`author'), however the first snippet relates to the first meaning, while the second snippet shows the third one:
\begin{enumerate}\foreignlanguage{russian}{
\item \textit{`МакSим (Марина Абросимова) – одна из самых популярных и коммерчески успешных певиц в России, являющаяся \textbf{автором} и исполнителем...'}
\item \textit{`Работает с 1994 года. Книги и тексты, разбитые по жанрам и \textbf{авторам}.'}}
\end{enumerate}
That means that there is a need for more sophisticated way to cluster search results. We should somehow learn which senses the query has and with which words these meanings are (probabilistically) associated. One of the possible ways to solve this problem is by extracting co-occurrence statistics from large corpora. The idea behind this is that word meaning is in fact the sum (or the average) of its uses. So, meaning is a function of distribution (cf. \cite{Harris:1970}). Thus, if we know with which words the query typically co-occurs and how these neighbors are related to each other, then we know the `sense set' of the query. After that we can somehow measure semantic similarity of each search snippet on the SERP with each of the senses and map them to each other. This information can then be used to either rank the results, or mark them with appropriate labels.

The structure of the paper is as follows. In Section \ref{section:related} we briefly overview work previously done on the subject. Section \ref{section:graph} describes the process of building co-occurrence graphs from large Russian corpora. In Sections \ref{sec:querygraph} and \ref{sec:senses} we conduct an experiment on clustering SERPs with ambiguous queries from Mail.ru search engine with the help of the methods described before. The results are evaluated in Section \ref{section:clustering_eval}. Section \ref{section:conclusion} draws conclusions concludes and provides suggestions for further research.

\section{Related Work}\label{section:related}
As stated in the previous Section, we are inspired by a fundamental hypothesis than meaning depends on the distribution \cite{Harris:1970} and that frequency of linguistic phenomena (in our case, word co-occurrence) is important for determining these phenomena's place in the system of language \cite{Bybee:2006}. Our work is also based on the idea that the senses of ambiguous lexical units should be induced from the data itself, not from a dictionary. No dictionary is perfect or comprehensive, because \textit{`senses as identified in the dictionary identify points on a continuum of possibilities for how the word is used'} \cite{Kilgariff:1992}. The only robust source of words' meanings in the text is the text itself. That's why we shift our focus away from selecting the most suitable senses from a pre-defined inventory towards discovering senses automatically from the raw data, which is natural text.

One of the first notes on practical application of this idea to word sense disambiguation and word sense induction is found in \cite{Schutze:1995}, where vector representations of word similarity derived from co-occurrence data are used. Broad review of contemporary (by 2012) state of the field is provided in \cite{Navigli:2012}. 

The main source of methods for our present research is \cite{Navigli:2013}, which describes workflow for clustering web search results using graph analysis over co-occurrence networks. Specifically, we use the notion of query graph, consisting of query terms and words from search engine results page augmented with nearest neighbors and relations from a reference corpus. For partitioning query graph and clustering query senses we employed \textit{Curvature} algorithm \cite{Navigli:2013} and \textit{Hyperlex} algorithm proposed in \cite{Veronis:2004}.

\section{Building Co-Occurrence Graph}\label{section:graph}

The first thing we had to do was to select a text corpus to build the graph upon. It is well known that the larger the corpus is the more co-occurrence information it contains. However, increasing corpus size also leads to exponentially growing computation time. Thus, for the sake of time and because of the preliminary nature of our research, we restricted ourselves to three Russian corpora of smaller but still decent size:
\begin{enumerate}
\item Open Corpora\footnote{\url{http://opencorpora.org}} (1 million tokens), further \textit{`OC'};
\item Disambiguated fragment of Russian National Corpus\footnote{\url{http://ruscorpora.ru}} (1 million tokens), further \textit{`RNC'};
\item Corpus of random search queries from Mail.ru search engine\footnote{\url{http://go.mail.ru}} (2 million tokens), further \textit{`QC'}.
\end{enumerate}
The first two items are academic corpora of Russian texts, supposedly representing (written) language in general. They differ in that the first one consists of full texts published under various free and open licenses, while the second one is a random sample of sentences from the larger Russian National Corpus. Both of them come with morphological annotation.

The third corpus was taken for comparison. It is important in view of the aim of our research (to test semantic SERP clustering). Our intuition was that perhaps query corpus provides more `real-life' sense inventory. It is two times as big as its counterparts, because `connectivity' between its members is lower (see Table \ref{tab:graphs}) and we had to compensate for this.

At the same time, it turned out that the first two corpora mixed into one give better results, thus below we will often refer to such `meta-corpus' as \textit{`Mix corpus'}.

Before constructing the graph itself, we preprocessed the corpora, namely:
\begin{enumerate}
\item Removed from \textit{QC} all queries which did not contain Cyrillic characters (as apparently they are not Russian), 
\item Processed \textit{QC} with Freeling analyzer \cite{Padro:2012} to extract lemmas and morphological information for all tokens,
\item Removed stop words,
\item Removed all tokens except nouns, as we restrict ourselves to inducing only nominal senses (the same strategy was applied in \cite{Navigli:2013}).
\end{enumerate}

Sizes of preprocessed corpora are given in Table \ref{tab:corpora}.
\begin{table}[h]
 \begin{center}
		\caption{Sizes of corpora participating in the experiment}\label{tab:corpora}
		\begin{tabular}{|l|r|}
 		\hline
		\textbf{Corpus}&\textbf{Size (tokens)}\\ \hline
		\textit{OC} &490671 \\ 
		\textit{RNC} & 294849 \\ 
		\textit{Mix} & 785520 \\ 
		\textit{QC} & 1035483 \\ \hline
		\end{tabular}
 \end{center}
\end{table}
Average query length in QC is 2.47 noun tokens per query. 

After the corpus has been built, the process of constructing co-occurrence graph is rather straightforward: we create an empty graph and then populate it with vertexes denoting word types in the text (lemmas). After that for each lemma we find all its immediate neighbors in the corpus, that is, words to the left and to the right (sentence boundaries not crossed, queries considered to be `sentences' as well). If two lemmas were neighbors at least one time, we draw an edge between corresponding neighbors. 

Finally, we have an undirected graph in which noun lemmas are vertexes and co-occurrence relations are edges. For each edge we also calculate Dice coefficient \cite{Smadja:1996}. It measures the `strength' of the collocation, based on absolute frequency (\textit{c}) of both words (\textit{w} and \textit{w'}) and collocation (\textit{w,w'}):
\begin{equation}\label{eq:dice}
Dice(w,w') = \frac{2c(w,w')}{c(w)+c(w')}
\end{equation}
One can also think about the graph as a matrix of Dice coefficient values for all possible pairs of lemmas in the corpus.

Table \ref{tab:graphs} gives an overview of the basic features of the graphs.

\begin{table}
		\caption{Parameters of the graphs}\label{tab:graphs}
		\begin{tabularx}{\textwidth}{|XXXXXX|}
 		\hline
		\textbf{Corpus} & \textbf{Vertexes} & \textbf{Edges} & \textbf{Average degree} & \textbf{Average path length} & \textbf{Clustering coefficient}\\ \hline
		\textit{OC} & 21881 & 257846 & 23.57 & 3.26 & 0.166 \\ 
		\textit{RNC} & 22467 & 163914 & 14.6 & 3.53 & 0.136 \\ 
		\textit{Mix} & 31984 & 395225 & 24.7 & 3.29 & 0.186 \\ 
		\textit{QC} & 85548 & 291033 & 6.8 & 4.07 & 0.16 \\ \hline\end{tabularx}
\end{table}
One can see that the average degree of \textit{QC} is lower in comparison with the other corpora (because queries are typically shorter that sentences in natural texts). That is one of the reasons for our decision to use a larger query corpus.

It should also be noted that all corpora comply to `small world' definition \cite{Watts:1998}, because their average path length is approximately the same as in a random graph with the same number of vertexes ($N_V$) and average degree ($A_D$), while clustering coefficient is significantly higher than it should be in the random graph. For example if \textit{Mix} corpus were a random one, its average path length would be equal to 3.24 ($=\frac{log(N_V)}{log(A_D)}$), very close to the actual value. However, in this case, its clustering coefficient should be 0.0015 ($=\frac{2\times A_D}{N_V}$), which is significantly lower than the actual value. The same is true for all other corpora.

`Small world' nature of our graphs means that vertexes in them tend to bundle into clusters, which is typical of many real-world networks. This finding supports the idea of extracting senses from such clusters. It also additionally proves the applicability of graph sense induction methods to our corpora, as English-language graphs in the related publications also showed such properties.

\section{Building Query Graph}\label{sec:querygraph}
We experimented with clustering search engine results page on a set of sixty ambiguous one-word Russian queries, taken from \textit{Analyzethis} homonymous queries analyzer\footnote{\url{http://analyzethis.ru/?analyzer=homonymous}}. \textit{Analyzethis} is a search engines evaluation initiative, offering various search performance analyzers, including one for ambiguous or homonymous queries. We crawled Mail.ru search for these queries, getting titles and snippets (10 for each result).

The procedure of semantic clustering starts with building the so called query graph. Here we closely follow \cite{Navigli:2013}.

First, we lemmatize all snippets and titles and remove stop words and the query word itself. Then we construct a graph $G_q$ with all nouns from snippets and titles as vertexes. Then we use one of the large corpora graphs (those that we built in Section \ref{section:graph}) to find words strongly connected to the query word and add these words to the query graph. We consider a connection `strong' if it falls under the following constraints:
\begin{equation}
	\left\{
		\begin{aligned}\label{eq:querygraphconstraints}
			\frac{c(q,w)}{c(q)} \ge 0.01\\
			Dice(q,w) \ge 0.005\\
		\end{aligned}
	\right.
\end{equation}
where \textit{c} is absolute frequency in the corpus, \textit{q} is the query and \textit{w} is the word under analysis. Thresholds 0.01 and 0.005 were determined empirically while experimenting on the above mentioned ambiguous queries set. These thresholds produced most convincing sense clustering. However, the issue of choosing the thresholds is a subject for thorough evaluation in future.

Thus, we now have $G_q$ with no edges and vertex set consisting of words from the search result and strong neighbors of the query word. After that, for each pair of words (\textit{w},\textit{w'}) in $G_q$ we check if they co-occur in the large corpus. If they do and $Dice(w,w')\ge0.005$, we connect these words in $G_q$ with an edge with weight = \textit{Dice(w,w')}. Finally, we delete disconnected vertexes (those with the degree equal to 0).

\section{Processing Query Senses and Results}\label{sec:senses}
With query graph at hand, we are ready to find which senses the query has. What we need is an optimal partition of the query graph, in which words related to different senses are in different parts of the graph. We apply two techniques for that, namely, \textit{Curvature} from \cite{Navigli:2013} and \textit{Hyperlex} from \cite{Veronis:2004}.

\begin{wrapfigure}{r}{0.6\textwidth}
\includegraphics[width=0.6\textwidth]{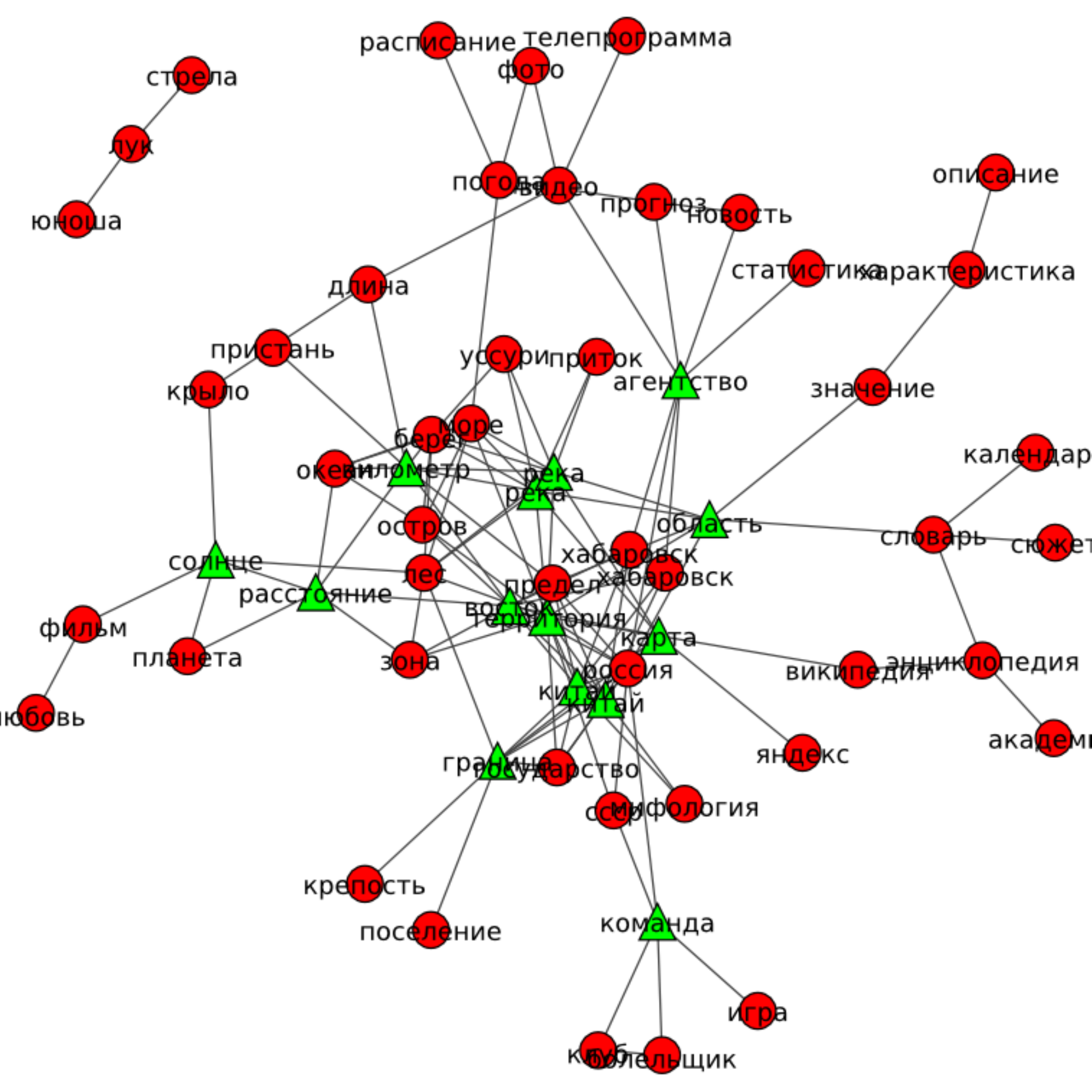}
\caption{Query graph for \foreignlanguage{russian}{\textit{`амур'}} (\textit{Curvature})}\label{fig:qg_amur}
\end{wrapfigure}

\subsection{Curvature}
\textit{Curvature} algorithm aims at finding vertexes from $G_q$ with low local clustering coefficient. Our hypothesis is that these are words which serve as `links' between different senses or `uses' of the query. Then we remove vertexes with clustering coefficient below a certain threshold. It leads to the graph disjointing into several components related to different senses. Vertexes in these components represent lexical inventory of each sense. Disconnected vertexes are removed from the final graph.

Let us illustrate the process with the example of \foreignlanguage{russian}{\textit{`амур'}} (`Amur') query. Figure \ref{fig:qg_amur} shows its query graph. It is already disconnected into two components and the meaning of love god (associated with words \foreignlanguage{russian}{\textit{`лук'}, \textit{`стрела'}} and \foreignlanguage{russian}{\textit{`юноша'}}) is separated. However, other `senses' of the query remain hidden in the giant component. Vertexes shown as triangles have low clustering coefficient and are thus marked for deletion.

So, we delete `triangular' vertexes. Note that we chose threshold 0.3 -- all vertexes with clustering coefficient below this are removed. It is also important that we do not delete vertexes with clustering coefficient = 0. This is because neighbors of such vertexes are not connected to anything except this vertex. If we remove it, a lot of disconnected vertexes will appear. Such clusters (consisting of only one word) do not make much sense. For example, the word \foreignlanguage{russian}{\textit{`лук'}} on Figure \ref{fig:qg_amur} is characterized by clustering coefficient = 0. If we remove it, then the whole component representing `love god' meaning disappears.

\begin{wrapfigure}{l}{0.6\textwidth}
\includegraphics[width=0.6\textwidth]{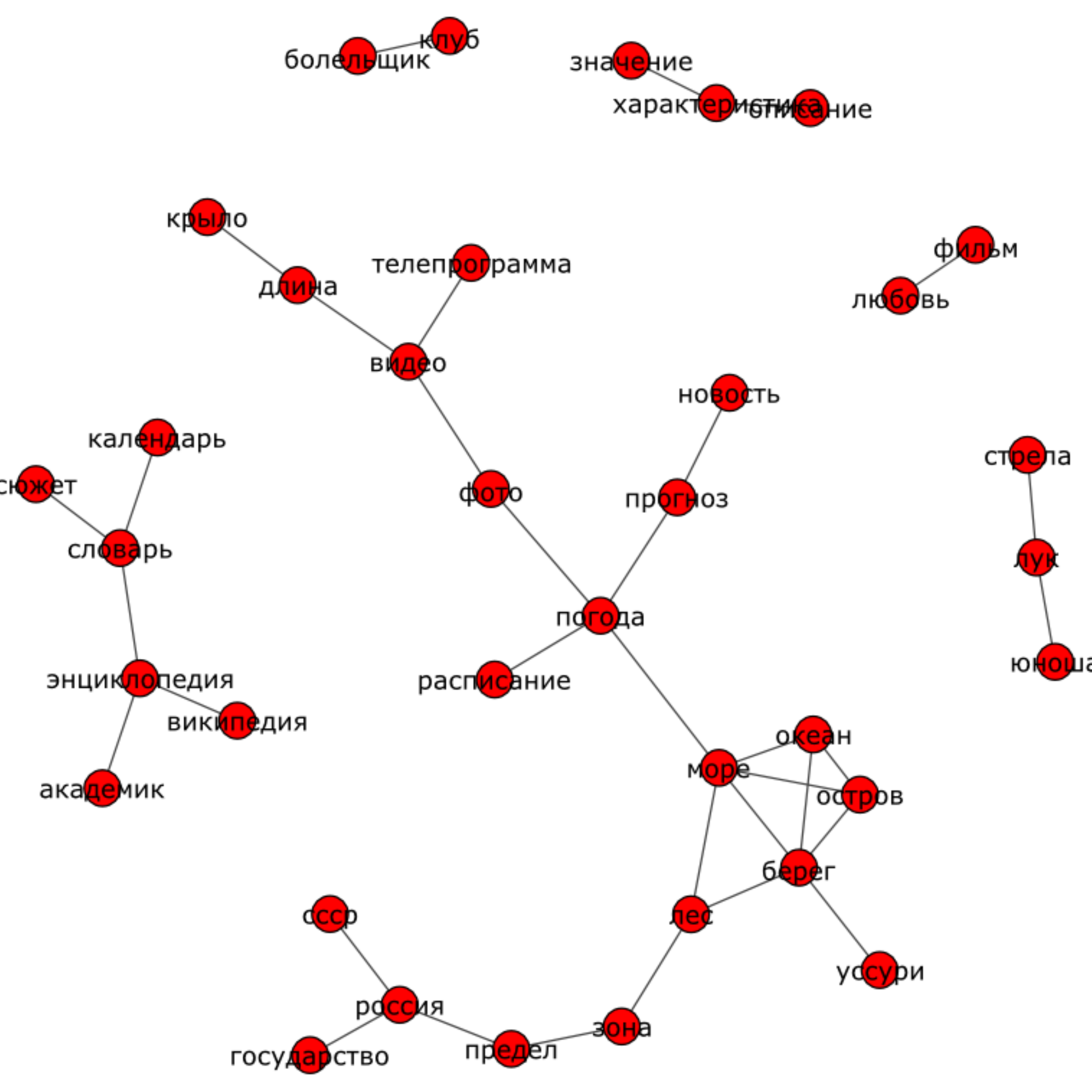}
\caption{Disjointed query graph (\textit{Curvature})}\label{fig:qg_amur2}
\end{wrapfigure}

Figure \ref{fig:qg_amur2} shows the query graph after removing vertexes with low clustering coefficient. We now have 6 components (note that the labels for these clusters are introduced by us, not by the algorithm):
\begin{enumerate}
\item River (all vertexes except enumerated below)
\item Love god \foreignlanguage{russian}{(\textit{`юноша, лук, стрела'})}
\item Hockey club \foreignlanguage{russian}{(\textit{`клуб, болельщик'})}
\item Movie \foreignlanguage{russian}{(\textit{`любовь, фильм'})}
\item Dictionary-1 \foreignlanguage{russian}{(\textit{`календарь, словарь, википедия, энциклопедия, академик, сюжет'})}
\item Dictionary-2 \foreignlanguage{russian}{(\textit{`значение, описание, характеристика'})}
\end{enumerate}

First 4 components clearly represent different meanings of the word \foreignlanguage{russian}{\textit{`амур'}}. The last two are rather `uses', typical contexts. However they can still be useful in clustering as they allow to keep encyclopedic results together.

\subsection{Hyperlex}
\textit{Hyperlex} algorithm described in \cite{Veronis:2004} introduces the notion of `hubs' within the graph, meaning most inter-connected vertexes and employs the graph's maximum spanning tree. Just like the previous algorithm, it takes as an input the query graph $G_q$ we prepared in Section \ref{sec:querygraph} and the query itself.

First we create a list \textbf{L} with all vertexes from $G_q$ sorted in decreasing order by their absolute frequency in the large corpus. Then for each item of this list we check if the corresponding vertex complies to the following constraints:
\begin{enumerate}
\item Vertex normalized degree is greater than or equal to 0.05,
\item Average Dice coefficient of vertex edges is greater than or equal to 0.007.
\end{enumerate}

If the constraints are met, we add this word to the hub list, considering it to be a kind of a connector. Simultaneously, we remove this vertex and its neighbors from the list \textbf{L} and continue iterating. In case we meet a word which does not satisfy the requirements above, we check whether the list of hubs has at least two elements. If it does, we stop iterating, if not, we continue to the next item. Note that it differs from the original \textit{Hyperlex} algorithm, where one should stop no matter how long the hub list is. In our Russian material it sometimes caused the hub list to remain empty or contain only one item, which is useless.

After we have the list of hubs, we augment $G_q$ with query vertex and connect this vertex to all hubs putting infinite (or very high) Dice coefficient on the corresponding edges. Then, we produce a maximum spanning tree from this graph. Maximum spanning tree is an attempt to keep all the vertexes connected while eliminating cycles and using as few edges as possible with as high weights (in our case it is Dice coefficient) on them as possible. In the spanning tree, there is only one path between any two vertexes and this path lies through edges with maximum Dice. Because the query vertex and the hubs are connected by edges with infinite Dice, they are sure to be the center of the spanning tree and directly linked.

\begin{wrapfigure}{r}{0.6\textwidth}
\includegraphics[width=0.6\textwidth]{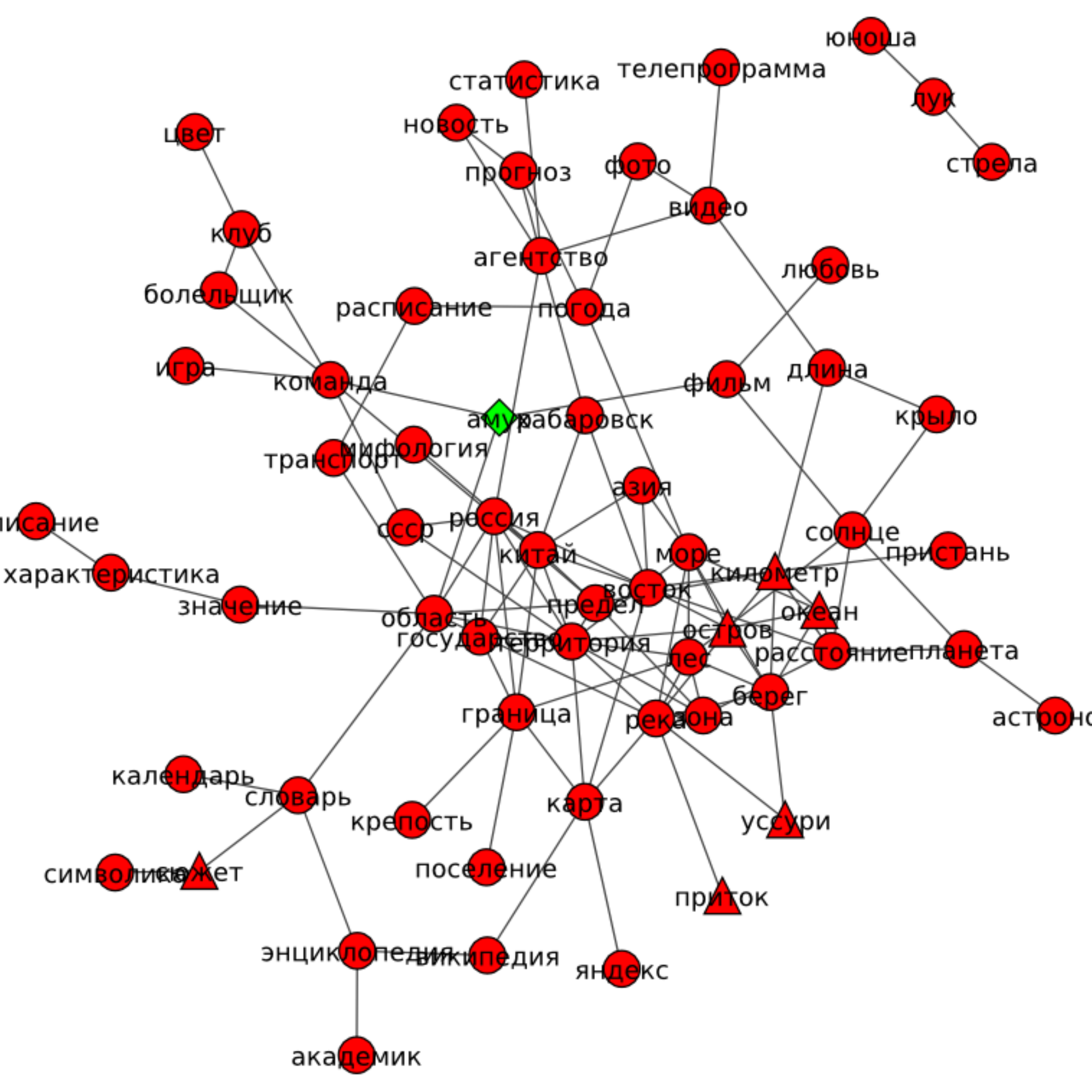}
\caption{Query graph for \foreignlanguage{russian}{\textit{`амур'}} after augmenting it with the query vertex (\textit{Hyperlex})}\label{fig:qg_stalker2}
\end{wrapfigure}

At last we remove the query vertex from the spanning tree, producing disjointed subtrees with hubs as roots. These subtrees represent query meanings. Note that we also delete all disconnected vertexes (those with degree = 0).

Let us present an example of Hyperlex at work with the same query \foreignlanguage{russian}{\textit{`амур'}}. Our corpus is \textit{Mix}. Initial state of the query graph $G_q$ is the same as in Figure \ref{fig:qg_amur}.

We add the word \foreignlanguage{russian}{\textit{`амур'}} to the graph $G_q$ and connect it to vertexes selected as hubs: \foreignlanguage{russian}{\textit{`область, фильм, команда'}}. The result is presented on Figure \ref{fig:qg_stalker2} with query vertex drawn as a diamond. For reference, vertexes which were introduced from the corpus and not from search results \foreignlanguage{russian}{(\textit{`сюжет'}, \textit{`океан'}}, etc) are drawn as triangles.

\begin{wrapfigure}{l}{0.6\textwidth}
\includegraphics[width=0.6\textwidth]{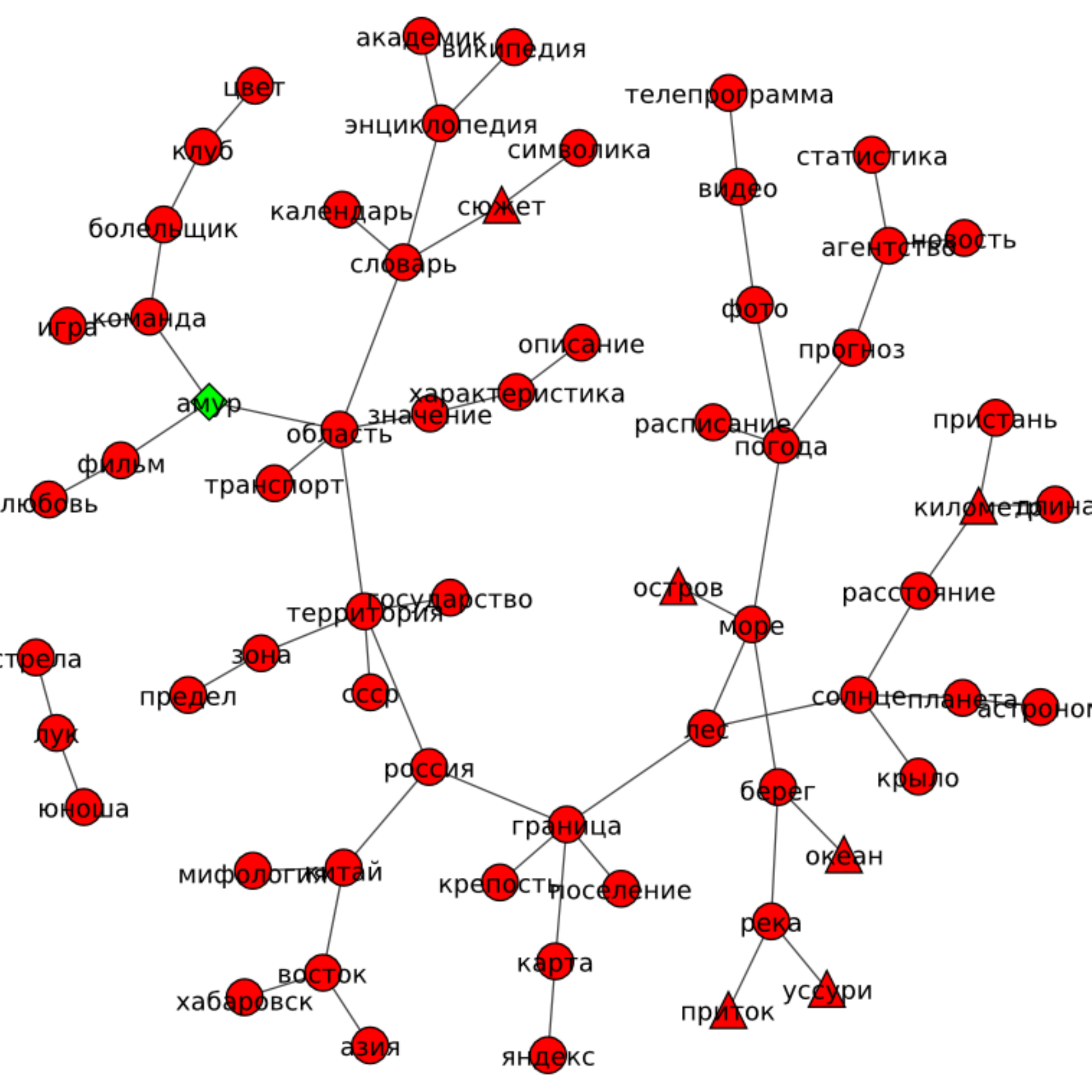}
\caption{Maximum spanning tree for \foreignlanguage{russian}{\textit{`амур'}} query graph (\textit{Hyperlex})}\label{fig:qg_stalker3}
\end{wrapfigure}

Now we produce maximum spanning tree from $G_q$ with Dice coefficient as weight measure. The tree is visualized on Figure \ref{fig:qg_stalker3}. Note that it has much fewer edges than the initial $G_q$.

\begin{wrapfigure}{r}{0.6\textwidth}
\includegraphics[width=0.6\textwidth]{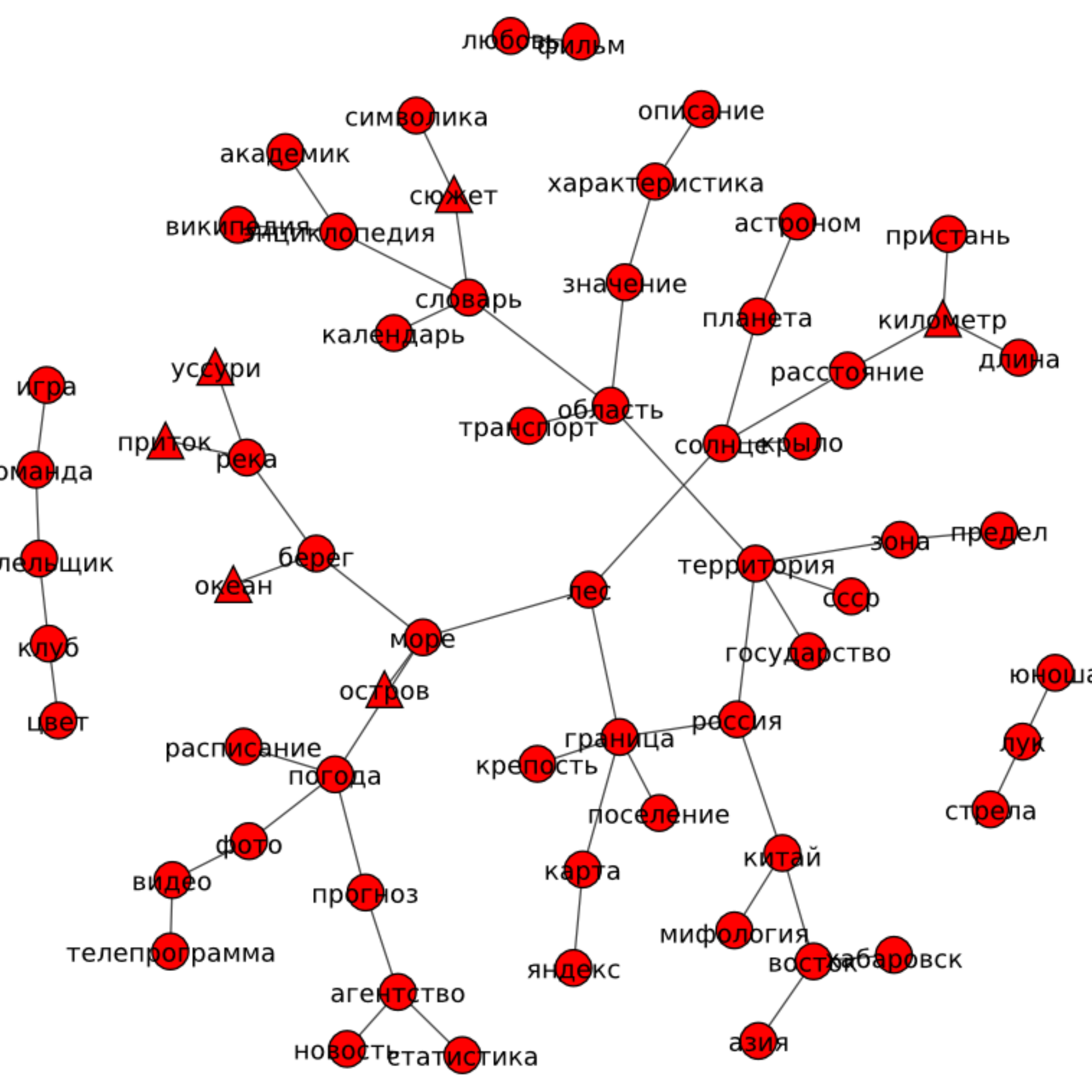}
\caption{Maximum spanning tree after removing query vertex (\textit{Hyperlex})}\label{fig:qg_stalker4}
\end{wrapfigure}

Finally, we remove the query vertex and all vertexes that become disconnected after this removal. As a result, we have a disjointed graph shown in Figure \ref{fig:qg_stalker4}. The number of the components has grown from 2 to 4 (once again, labels are assigned by us):

\begin{enumerate}
\item Love god \foreignlanguage{russian}{(\textit{`юноша, лук, стрела'})}
\item Movie \foreignlanguage{russian}{(\textit{`любовь, фильм'})}
\item Hockey club \foreignlanguage{russian}{(\textit{`клуб, игра, болельщик, команда, цвет'})}
\item River (all the remaining vertexes)
\end{enumerate}

One can see that \textit{Hyperlex} successfully extracted the same four important meanings as the previous algorithm. At the same time, unlike \textit{Curvature}, it managed to avoid two `encyclopedic' clusters (obviously in common for too many queries) and leave their vertexes in the `river' cluster. Also, \textit{Hyperlex} is better because it describes `hockey club' cluster in a richer way, using 5 relevant words instead of 2.

One can again note that in fact what we call `senses' are not senses like meanings in the dictionaries. We agree with Jean Veronis who argues that co-occurrence networks reflect `uses' rather than senses. So, what we have are typical environments where the word is used, and these environments are only loosely connected to what a lexicographer would call `senses' or `meanings'. However, we are fine with that, as we assume that clustering SERP according to `typical uses' is at least equally important as clustering according to `proper senses'. Perhaps, these senses are in fact less related to real-life, as even linguists sometimes have trouble matching the `senses' found in a dictionary and the occurrences found in a corpus \cite{Veronis:2004}. Additionally, as has already been stated, dictionary senses are always limited and by design cannot cover new semantic trends and subtle meanings quickly appearing and disappearing in the modern world. Thus, theoretically typical uses are more relevant for clustering than academic dictionary senses. To strictly prove it for the Russian material, one needs manually clustered data set (see Section \ref{section:clustering_eval}), and we leave it for further research.

\subsection{Mapping Results to Senses}\label{section:mapping}
Once we possess the sense inventory for the query, we can combine it with bags-of-words for each search result to finally perform SERP clustering. We do that in a rather straightforward way.

Given a set of senses represented by a lemma set each and a set of results (snippet and title) represented by lemma sets as well, for each pair of result ($r$) and sense ($s$) we calculate similarity measure $sim$. It is a simple number of lemmas in common for both sets divided by the number of lemmas in the result:
\begin{equation}
sim(r,s) = \frac{r\cap s}{length(r)} \label{eq:sim}
\end{equation}

Then we choose the sense with maximum similarity and link this sense to the result. Thus, each result receives some sense, and is `understood'. 

In the future we plan to explore other means of calculating similarity measure as well, for example, counting tokens not types or considering weights on edges in the intersection.

\section{Evaluation of SERP clustering}\label{section:clustering_eval}

Generally, evaluation of clustering is a rather harsh task. Perhaps, the best way to do this is to employ human assessment, but for the time being we limited ourselves to simple evaluation of the correctness of cluster number (that is, number of meanings). 

\textit{Analyzethis} service provides data about how many senses of an ambiguous query are there in the SERP. Thus we consider it to be an expert opinion and check how strong is our deviation from this `gold standard'. For example, if \textit{Analyzethis} believes that there are three senses present on the SERP, and our clustering algorithm puts all the results into one cluster, this signals that the algorithm is not optimal. The same is true if the number of clusters is, for example, eight. The less our deviation from \textit{Analyzethis} assessment is, the better. So, in fact we check that the employed algorithms do not produce senseless results (too many or too few meanings). We once again note that in order to evaluate the contents of the clusters themselves, one needs manually clustered SERPs for ambiguous queries. To our knowledge, there is no such a data set for Russian. We are working on creating it.

For the time being, we compared the number of clusters for each of ambiguous queries in four different settings (two corpora and two word sense induction methods). Then we calculated average deviation of our clustering number from that of \textit{Analyzethis}. Table \ref{tab:evaluation} provides the results of this comparison. Note that the average number of senses per query in \textit{Analyzethis} data set was 2.65. 

\begin{table}
 \begin{center}
		\caption{Evaluation of SERP clustering (average deviation from \textit{Analyzethis} assessment in number of senses and in percent from the average number of senses in the set)}\label{tab:evaluation}
		\begin{tabular}{|c|cc|}
 		\hline
		\textbf{Corpus}& \textbf{Curvature} & \textbf{Hyperlex} \\ \hline
		\textit{Mix} & 1.636 (61\%) & \textbf{1.288} (49\%) \\ 
		\textit{Query} & 1.742 (66\%) & 1.379 (52\%) \\ \hline
		\end{tabular}
 \end{center}
\end{table}
It is clear that \textit{Hyperlex} consistently outperforms \textit{Curvature}, and that \textit{Mix} corpus does the same with the query corpus. \textit{Hyperlex} victory comes as no surprise, as it uses maximum spanning tree notion, which seems to allow deeper grasping of graph structure. The victory of \textit{Mix} corpus (which is smaller than the query corpus) is much less expected. We believe that there are two reasons for this:

\begin{enumerate}
\item As we have already mentioned, the query corpus is less `dense' because of low length of queries. Thus, there are fewer edges and less data for algorithms.
\item Query corpus was lemmatized with Freeling while \textit{Mix} corpus consists of manually annotated corpora. Glitches and outright errors of Freeling could impact graph quality. This can be fixed in the future either by improving Freeling or by using another lemmatizer.
\end{enumerate}
Thus, at the moment, using \textit{Mix} corpus and \textit{Hyperlex} algorithm of word sense induction seems to be the best option. However, things surely can be different if we employ larger corpora (which we plan to do in the future).

\section{Conclusion and future work}\label{section:conclusion}
We showed that state-of-the-art methods of word sense induction and search results clustering based on semantic graphs do work for Russian data. 

Application of such methods can lead to search engine results presentation getting closer to actual semantics of the results, not simply term frequency ranking. For a user, it would mean the possibility to immediately grasp which results in the SERP are actually related to the query sense, and which other senses exist. The power of this approach can be increased by wider employment of Semantic Web paradigm: semantically marked up web pages are represented by generally better and clearer snippets. Such snippets, in turn, should provide better data for graph-based word sense induction algorithms. 

We plan to experiment on more types of query graph processing and launch a full-scale human evaluation of results. Also, it seems profitable to use not only separate words, but also compound phrases, as well as to construct graphs with not only immediate neighbors, but also with second-order co-occurrences (neighbors of neighbors). Additionally, experiments with larger query corpora may lead to new and inspiring insights in this field.

\bibliographystyle{splncs}
\bibliography{reference}

\end{document}